\documentclass[runningheads]{llncs}

\usepackage{eccv}

\usepackage{eccvabbrv}
\usepackage{todonotes}
\usepackage{graphicx}
\usepackage{booktabs}

\usepackage[accsupp]{axessibility}  % Improves PDF readability for those with disabilities.

\usepackage{hyperref}

\usepackage{orcidlink}

\DeclareRobustCommand{\insaitlogo}{%
  \raisebox{-0.15ex}{\includegraphics[height=1.5ex]{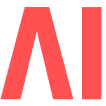}}%
}

\usepackage{marvosym}

\DeclareRobustCommand{\adobemark}{\ensuremath{\lozenge}}

\begin{document}

\title{Knowledge-Centric Agents for Workflow Generation in ComfyUI}

\titlerunning{Knowledge-Centric Agents for Workflow Generation}

% Include the authors' ORCID for the camera-ready version, if at all possible.
% \author{
% Zhendong Li~\textsuperscript{\insaitlogo} \and
% Lei Sun~\textsuperscript{\insaitlogo} \and
% Ruibo Ming~\textsuperscript{\insaitlogo} \and
% He Zhang~\textsuperscript{\adobemark} \and
% Danda Pani Paudel~\textsuperscript{\insaitlogo} \and
% Luc Van Gool~\textsuperscript{\insaitlogo} \and
% Jinjin Gu~\textsuperscript{\insaitlogo}
% }

% \institute{ \textsuperscript{\insaitlogo} INSAIT, Sofia University ``St. Kliment Ohridski'', \quad \textsuperscript{\adobemark} Adobe Research
% \texttt{\{zhendong.li, lei.sun, ruibo.ming, danda.paudel, luc.vangool, jinjin.gu\}@insait.ai, hezhan@adobe.com}
% }

\author{
Zhendong Li~\textsuperscript{\insaitlogo} \and
Lei Sun~\textsuperscript{\insaitlogo,\Letter} \and
Ruibo Ming~\textsuperscript{\insaitlogo} \and
He Zhang~\textsuperscript{\adobemark} \and
Danda Pani Paudel~\textsuperscript{\insaitlogo} \and
Luc Van Gool~\textsuperscript{\insaitlogo} \and
Jinjin Gu~\textsuperscript{\insaitlogo,\Letter}
}

\institute{
\textsuperscript{\insaitlogo} INSAIT, Sofia University ``St. Kliment Ohridski'',
\quad
\textsuperscript{\adobemark} Adobe Research
\texttt{\{zhendong.li, lei.sun, ruibo.ming, danda.paudel, luc.vangool, jinjin.gu\}@insait.ai, hezhan@adobe.com}
\\
\Letter~\textit{Corresponding authors.}
}

\authorrunning{Z.~Li et al.}

\maketitle
\begin{center}
    \centering
    %\captionsetup{type=figure}
    \captionsetup{type=figure}
    \includegraphics[width=\linewidth]{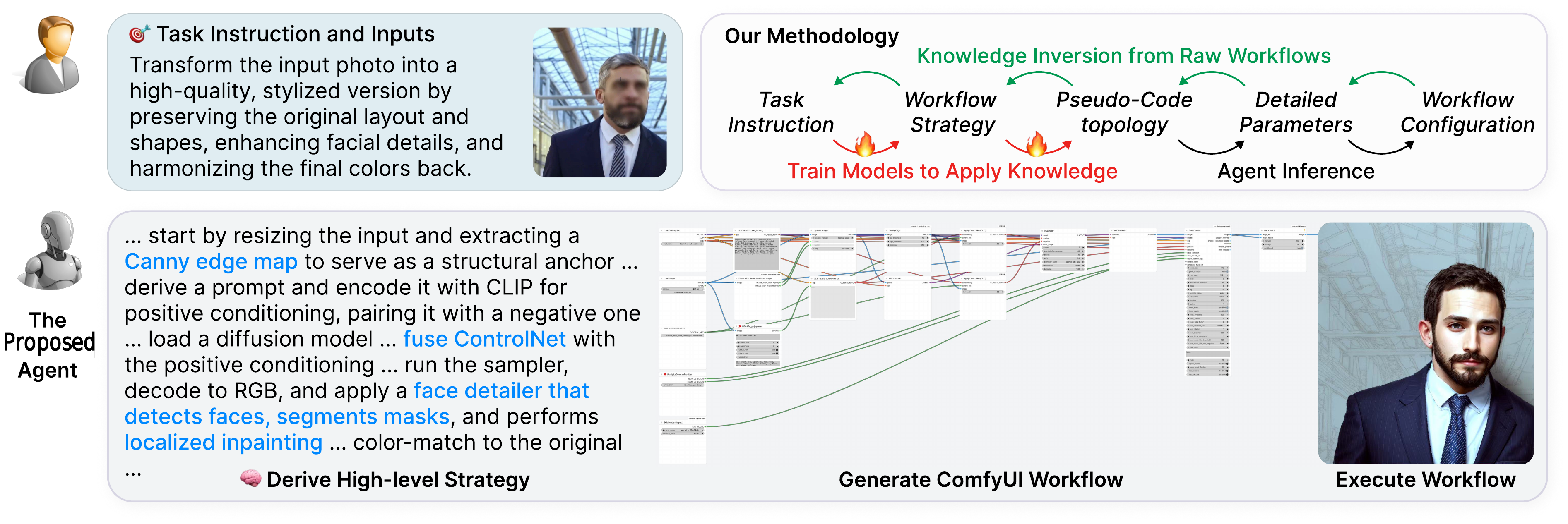}
    \captionof{figure}{Given a user task instruction, our agent learns to construct and execute a complete ComfyUI workflow. Through knowledge inversion, it distills pseudo-code structures and workflow strategies from large collections of real workflows. During inference, During inference, the agent executes the generated workflow to produce a high-quality visual result. The example shows how the agent interprets a stylistic editing request, derives a high-level strategy, composes the corresponding ComfyUI graph, and executes it to generate the final output.}
    \label{fig:teaser}
\end{center}

\begin{abstract}
Workflow generation in visual creation systems such as ComfyUI demands not only syntactic accuracy but also expert-level reasoning over modular compositions.
Existing large language model (LLM) approaches often treat this as a direct text-to-JSON generation task, struggling with structural brittleness and lacking the experiential knowledge required for effective design.
We argue that successful workflow generation requires modeling knowledge itself, including its structure, hierarchy, and reasoning dynamics.
To this end, we propose a knowledge-centric framework that learns to \textbf{invert}, \textbf{inject}, and \textbf{infer with knowledge} across multiple abstraction levels.
We first perform knowledge inversion to distill hierarchical representations, ranging from full pseudo-codes and skeletons to high-level strategies, from large collections of real-world workflows.
We then conduct knowledge injection through supervised fine-tuning, teaching the model to reason from task descriptions to strategies and from strategies to executable structures.
During inference, the model performs reversible reasoning to synthesize executable workflows, augmented by self-refinement for structural coherence.
Extensive experiments demonstrate that our method produces workflows with richer node diversity, more coherent structures, and higher execution success rates than existing systems, establishing a new foundation for knowledge-driven, agentic workflow generation.
\end{abstract}

\section{Introduction}
\label{sec:intro}

The whirlwind of progress in modern computer vision models has steadily transformed controllable image generation, creation, editing, and processing from laboratory prototypes into powerful tools capable of addressing complex real-world demands.
In practice, both technical experts and artists often need to combine and coordinate multiple model components to accomplish those tasks.
This results in a non-monolithic, modular composition paradigm for problem solving, where different modules are orchestrated into a Directed Acyclic Graph (DAG), commonly referred to as a workflow, and executed sequentially or conditionally.
ComfyUI~\cite{comfyanonymous2023comfyui} embodies this philosophy and has become a widely adopted visual programming framework for building such workflows in image creation pipelines.
Developing intelligent agents that can automatically generate and refine these workflows is thus not only of great practical relevance but also poses deep scientific challenges.

A ComfyUI workflow can be fully represented as a JSON configuration, which specifies all nodes, parameters, and interconnections.
Several recent studies have attempted to leverage large language models (LLMs) to generate these configurations automatically~\cite{xue2025comfybench,gal2024comfygen,xu2025comfyuir1,guo2025comfymind,huang2025comfygpt,sobania2024comfygi}.
However, generating valid and effective workflows differs fundamentally from conventional code or text generation.
It requires multi-level reasoning that spans from abstract conceptual design to concrete syntactic realization, each demanding distinct forms of knowledge and representation.

At the strategic level, workflow design is an open-ended and experience-driven process rather than a deterministic programming task.
ComfyUI provides a rich ecosystem of node types and model variants, each exhibiting complex, often nonlinear behaviors.
Their combinations yield an immense compositional space, where small structural changes can produce radically different outcomes.
Human designers rely heavily on tacit expertise, iterative experimentation, and intuition to determine whether segmentation is needed, how to arrange processing branches, or which control modules should interact.
Such heuristic reasoning, grounded in experiential knowledge, remains highly challenging for LLMs to acquire or reproduce purely from textual data.

At the structural level, even when the high-level strategy is known, translating it into a syntactically correct and executable JSON configuration poses significant challenges.
Workflows are large discrete graphs with intricate input–output dependencies.
The agent must not only infer the correct topology but also instantiate valid node parameters and maintain type-safe data flow.
The brittleness of JSON syntax, coupled with the combinatorial explosion of possible graph structures, makes direct text-to-configuration generation highly unreliable.
Small mistakes such as misconnected links or inconsistent parameter bindings can easily make the workflow fail to execute.

% These challenges collectively reveal a deeper limitation: \textit{existing language models lack structured, multi-level knowledge about workflow composition}.
% %
% While they excel at surface-level pattern completion, they struggle to reason about the hierarchical dependencies, causal logic, and experiential heuristics that underlie expert workflow design.
% %
% To overcome this gap, we propose a knowledge-centric framework that explicitly reconstructs, injects, and utilizes such multi-level representations.

These challenges collectively reveal a deeper limitation: \textit{existing language models lack structured, multi-level knowledge about workflow composition}.
While they excel at surface-level pattern completion or straightforward task decomposition, they struggle to reason about the hierarchical dependencies, causal logic, and experiential heuristics that underlie expert workflow design.
Furthermore, directly annotating such tacit expertise at scale is highly impractical.
To overcome this gap, rather than proposing a standard generation pipeline, our main contribution is a \textbf{knowledge-centric paradigm}.
The core insight is \emph{Knowledge Inversion}: starting from raw, unannotated workflows, we recover the missing domain knowledge and subsequently reuse it to empower LLMs for CoT-style decision-making.

Our approach begins by \textbf{inverting knowledge} from large collections of real workflows to extract hierarchical representations of strategy, structure, and reasoning;
then \textbf{injects} this distilled knowledge into language models through supervised fine-tuning;
and finally \textbf{infers with knowledge}, performing reversible reasoning from task descriptions to executable workflows.
This unified paradigm transforms workflow generation from an ad-hoc text generation problem into a principled process of knowledge reasoning, paving the way for agentic systems capable of composing complex visual pipelines with expert-level reliability.
In practice, we curate a collection of high-quality ComfyUI workflows to serve as both training and evaluation data.
Experimental results demonstrate that our method produces workflows with richer node diversity, more complex yet coherent structures, and more accurate connectivity.
Compared with existing approaches, it achieves a higher proportion of successfully solved tasks.
\section{Related Work}
\paragraph{ComfyUI and Automatic ComfyUI Assistant.}
%%% simply introduce comfyui
As AI-generated content (AIGC) continues to advance~\cite{NEURIPS2020_4c5bcfec_diffusion,NEURIPS2021_49ad23d1_diffusion}, an increasing number of models and tasks, including text-to-image generation, controllable synthesis, and image editing, are emerging~\cite{kumari2023multi,ruiz2023dreambooth, li2023dreamedit, zhang2023adding}. This shift has moved the community from relying on a single model to composing multiple models to accomplish more complex and customized workflows. Consequently, ComfyUI~\cite{comfyanonymous2023comfyui} has become increasingly popular for its flexible, node-based design. ComfyUI connects modular function nodes into directed acyclic workflows, enabling users to build and reuse complex AIGC pipelines. However, these workflows can quickly become complex, often involving nodes for prompt refinement, foreground–background segmentation, super-resolution, and many other operations, which creates a strong demand for automated workflow design tools.
To address this need, a variety of automated workflow-design tools have emerged~\cite{gal2024comfygen,xu2025comfyuicopilot,sobania2024comfygi,chen2025symbolic,huang2025comfygpt,xu2025comfyuir1,xue2025comfybench,guo2025comfymind}. ComfyUI-Copilot\cite{xu2025comfyuicopilot} assists users in selecting appropriate parameters and models; ComfyGen\cite{gal2024comfygen} focuses specifically on text-to-image generation; ComfyBench\cite{xue2025comfybench} and ComfyMind\cite{guo2025comfymind} adopt template-based composition strategies that cover most common tasks, with the latter offering tree-structured rollback. Meanwhile, ComfyGPT\cite{huang2025comfygpt} and ComfyUI-R1\cite{xu2025comfyuir1} leverage supervised fine-tuning and GRPO~\cite{shao2024deepseekmath} reinforcement learning to inject workflow-design knowledge into LLMs. However, when new workflows emerge from the community, these methods must be retrained, limiting their flexibility and scalability for future extensions.

\paragraph{LLMs and LLM-Agent.}
%% LLM
LLMs have rapidly advanced natural language understanding and generation. BERT~\cite{devlin2019bert} introduced bidirectional masked language modeling, greatly improving contextual representation learning. The GPT series \cite{brown2020language} shifted toward large-scale autoregressive pretraining, enabling few-shot and in-context learning. Later, ChatGPT and GPT-4 \cite{achiam2023gpt} combined instruction tuning and RLHF to enhance alignment and interaction quality. Recent models such as LLaMA~\cite{touvron2023llama}, Qwen~\cite{yang2025qwen3}, Gemini \cite{team2023gemini,comanici2025gemini} further expand reasoning, efficiency, and multimodal capabilities. 
%% agent
Building on these foundations, LLM-powered agents have emerged as autonomous entities capable of perceiving their environments, planning over extended horizons, and interacting with diverse external tools~\cite{franklin1996agent,achiam2023gpt}. 
Compared with traditional symbolic or RL-based agents~\cite{silver2018general,hwangbo2019learning}, those driven by LLMs exhibit superior adaptability, compositional reasoning, and task generalization~\cite{chen2023autoagents,li2024multimodal}. Recent research has further enhanced their autonomy through graph-based reasoning~\cite{yao2024tree,besta2024graph}, dynamic tool utilization~\cite{qian2025toolrl,li2025torlscalingtoolintegratedrl,shen2024hugginggpt,wu2023visual}, and self-reflective learning mechanisms~\cite{shinn2024reflexion}.
Some studies further explore automated optimization of agentic workflows. For example, GPTSwarm~\cite{zhuge2024gptswarm} models language agents as nodes in an optimizable graph, while TextGrad~\cite{yuksekgonul2024textgrad} introduces gradient-like optimization to iteratively refine prompts and agent behaviors. In addition, recent work investigates scalable evaluation strategies for agent systems~\cite{zhuge2024agent}, while~\cite{chen2024humans} highlights potential biases in automated evaluation.
These LLM-agents are applied widely in the areas of code generation~\cite{cursor2024}, health care~\cite{luo2022biogpt,singhal2023large}, and scientific discovery~\cite{taylor2022galactica}. Moreover, as a challenging and complex problem, workflow generation is also well-suited to be addressed by LLM agents, which can decompose tasks, plan multi-step actions, and coordinate across tools.

\begin{figure*}[t]
    \centering
    \includegraphics[width=0.98\linewidth]{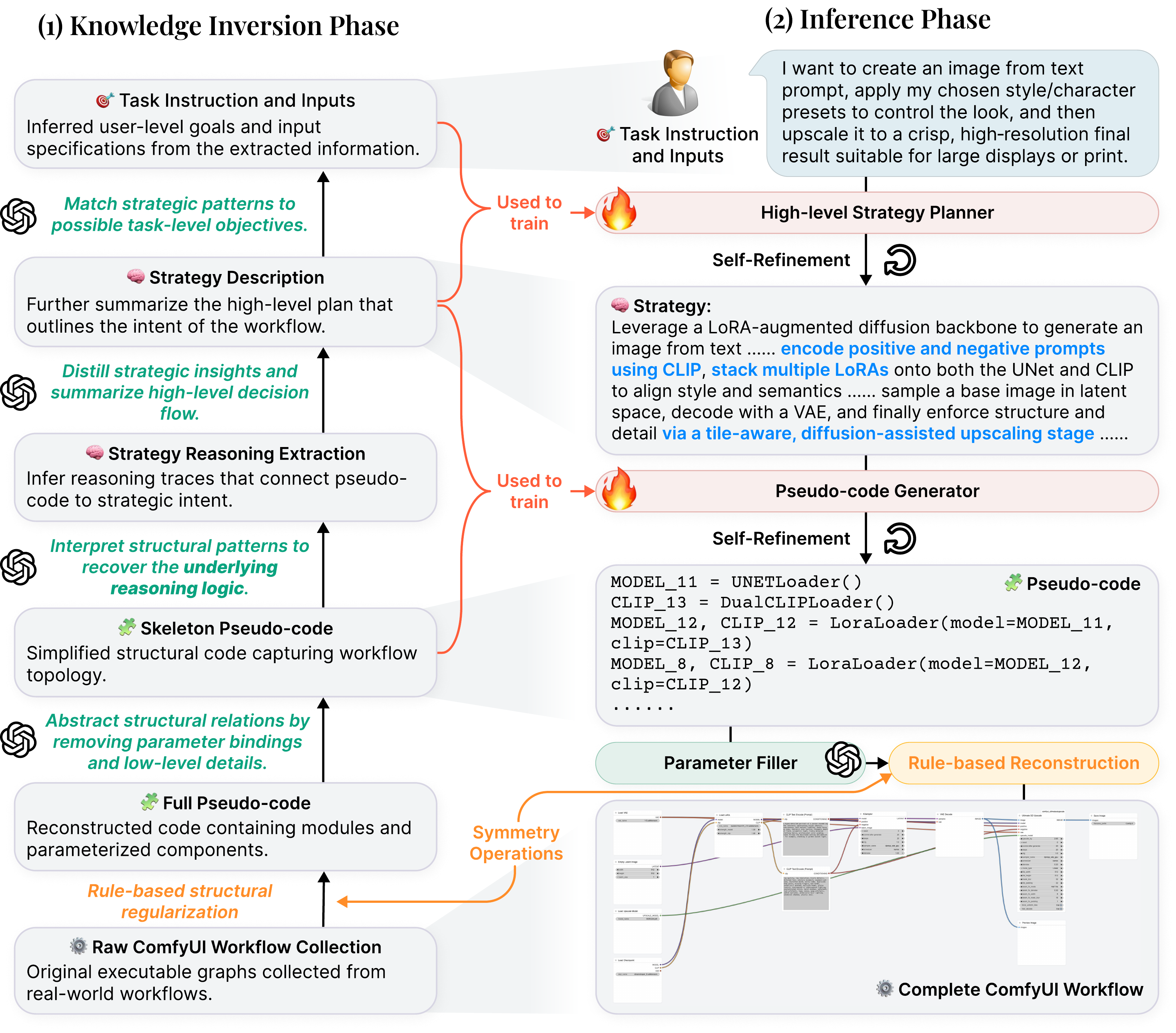}
    \caption{Overview of our framework.
The system consists of two symmetric phases.
(1) Knowledge Inversion distills hierarchical representations from real ComfyUI workflows.
(2) Knowledge Inference reverses this process to generate workflows from instructions via a strategy planner, pseudo-code generator, and rule-based reconstruction.}
    \label{fig:method}
\end{figure*}
\section{Method}

\subsection{Overview}
Generating a ComfyUI workflow is a complex problem with several layers of reasoning and representation.
A workflow is encoded as a JSON graph where nodes are functional modules and edges describe data or control dependencies.
Hence, workflow generation is equivalent to producing a valid and executable JSON file, which is far more delicate than ordinary text or code generation.

This complexity arises across several levels. Human experts do not jump from a task instruction directly to JSON.
They first form a high level strategy that decides the overall processing plan.
Typical decisions include whether segmentation is required, whether branches should process content separately, how many controls to introduce, and how controls interact.
This high-level strategy requires substantial domain knowledge and intuitive understanding of visual models and their compositional logic.

Given the strategy, experts construct the workflow topology, that is, a structured connection of modules.
This step is also difficult because the graph can be very large, discrete, and complex.
Directly generating JSON connections from text is extremely hard for language models due to syntactic brittleness and combinatorial explosion.
After the topology is formed, parameter tuning and detailed module configurations can be carried out, which are comparatively more rule-driven or pattern-based.
Therefore, workflow generation is a multi-level process, where each level poses distinct reasoning and learning challenges.
For example, translating topology into a full workflow can be rule-based, but deriving the topology itself from task instructions and strategies demands substantial knowledge and reasoning ability.

The key idea of our approach is to inverse knowledge, inject knowledge, and infer with knowledge across these hierarchical levels.
We systematically learn from large collections of real-world workflows, distill the underlying knowledge representations, and integrate them into a reasoning-capable agent system.

\subsection{Knowledge Inversion}
The goal of this stage is to inverse raw and heterogeneous workflow data into a set of hierarchical knowledge representations that can be effectively learned by language models or agent systems.
As illustrated in Figure~\ref{fig:method} (1), the entire process proceeds from the bottom to the top, progressively abstracting and reorganizing knowledge contained in large collections of raw workflows.

\begin{itemize}
    \item \textbf{From raw workflow to full pseudo-code.} We first employ a \textit{rule-based regularization} procedure to convert each workflow into a compact, machine-readable format called \textit{Full Pseudo-code}. This representation preserves all module names, topological connections, and parameter values, while removing the syntactic and ComfyUI-specific styling information.  This step does not involve any learning and serves as the structural foundation for higher-level abstraction.
    \item \textbf{To skeleton pseudo-code.} The next step abstracts the low-level implementation details from the Full Pseudo-code. A language model removes parameter bindings and numerical values while maintaining the functional topology of the workflow. The result is a \textit{Skeleton Pseudo-code} that captures only the essential graph structure and execution flow, providing a clean target for structure-aware learning.
    \item \textbf{To strategy reasoning.} Given the abstracted skeleton, we use large language models to perform semantic interpretation to infer the underlying reasoning traces that explain \emph{why} the workflow is organized in such a way. This \textit{Strategy Reasoning Extraction} step reconstructs the author's decision-making process. For instance, why multiple controls are fused, why branches are separated, or why a face-detailing module is placed after denoising. The output is a structured reasoning trace that links topological motifs to functional purposes.
    \item \textbf{To high-level strategy.} The inferred reasoning trace is then summarized into a concise \textit{Strategy Description}, which represents the high-level processing plan that an expert might verbally describe. This level captures the overall solution logic of the workflow, such as whether segmentation is required, which modules should cooperate, and how controls interact.
    \item \textbf{To task instruction and inputs.} Since most collected workflows lack explicit user prompts, we further infer the corresponding \textit{task instruction and input specifications} required for training. This is achieved by aligning the generated high-level strategy with retrieved exemplars and by decoding plausible task descriptions consistent with the structural and semantic context of the workflow. The result is a natural-language-level supervision signal that reconstructs the user intent behind each workflow.
\end{itemize}

This hierarchical inversion provides structured knowledge that can later be injected into large language models and utilized by agentic reasoning frameworks for workflow synthesis.
Each level can serve as either a hard target or a weak supervision signal for downstream training.

\subsection{Knowledge Injection}
After the hierarchical knowledge representations are distilled through inversion, the next stage is to inject them into language models so that the models can utilize this knowledge during inference.
Not all levels require knowledge injection.
Some transformations, such as filling in detailed parameters or expanding a skeleton pseudo-code into a complete version, can be handled by existing model priors or simple rule-based completion.
However, transitions that involve reasoning, creativity and experiments, such as mapping from a task description to a strategy, or from a strategy to a new pseudo-code, demand explicit knowledge injection.
We therefore categorize the knowledge injection process according to its learning necessity:
\begin{itemize}
\item \textbf{Task $\rightarrow$ Strategy.}
This transition encodes the experiential knowledge of human experts, illustrating how they analyze a task, decompose it into subgoals, and select processing modules accordingly.
Since this reasoning involves tacit expertise rather than explicit syntax, it cannot be inferred reliably by general-purpose models.
Supervised fine-tuning (SFT) is therefore essential to inject this experience-driven knowledge.

\item \textbf{Strategy $\rightarrow$ Skeleton Pseudo-code.}  
Translating a high-level plan into executable structures requires compositional reasoning and graph construction skills that the base model lacks.  
We again employ SFT to teach the model to generate structured, syntactically valid pseudo-codes consistent with the strategy’s intent.  
This step enables the model to internalize topological patterns and causal relations between modules.

\item \textbf{Skeleton $\rightarrow$ Full Pseudo-code.}  
This level mainly concerns parameter restoration and stylistic completion, which can be achieved through rule-based expansion or lightweight prompting.  
It does not require additional supervised training.

\end{itemize}

We emphasize that SFT is the primary mechanism for knowledge injection, as it directly teaches the model to produce new reasoning patterns or representations beyond its pretraining scope.
By contrast, reinforcement-style alignment methods such as GRPO are mainly effective for local preference optimization within an already feasible generation space.
In our setting, where the goal is to endow the model with new capabilities such as planning, decomposition, and workflow construction, SFT serves as the primary knowledge carrier.
Some studies indeed demonstrate that in similar settings, SFT continues to play the dominant role for capability injection \cite{xu2025comfyuir1}.

In practice, we fine-tune a large language model, Qwen3-14B ~\cite{yang2025qwen3}, on paired examples of \textit{(task, strategy)} and \textit{(strategy, pseudo-code)}.
During training, we also introduce an auxiliary objective of \textit{strategy reasoning extraction}, which requires the model to generate intermediate reasoning traces explaining each strategic decision.
This auxiliary task regularizes the model to learn the discourse and logical structure of expert reasoning, thereby improving both interpretability and generalization during workflow synthesis.

\subsection{Knowledge Inference}
Once the language model has been equipped with the injected knowledge, inference on new tasks becomes essentially the reverse process of knowledge inversion, as illustrated in Figure~\ref{fig:method} (2).
Given a user-provided task instruction and inputs, the system first invokes a \textit{High-Level Structure Planner}, an SFT-trained model that performs strategic reasoning to produce a high-level strategy describing the overall processing logic, decomposition, and control structure of the workflow.
The strategy reflects the distilled experiential knowledge obtained from the large corpus of real workflows.
Next, the generated strategy is passed to a \textit{Pseudo-code Generator}, another SFT-trained model that converts the abstract plan into an executable skeleton pseudo-code.
A pretrained model is then employed to fill in parameters and bindings based on the contextual semantics of each module, restoring a full pseudo-code with appropriate configuration values.
Finally, the rule-based reconstruction procedure converts the full pseudo-code into a fully functional ComfyUI workflow, which can be directly executed or visualized in the target system.

To further improve robustness, we incorporate self-refinement mechanisms after both the strategy planning and pseudo-code generation stages.
The model is prompted to re-evaluate and refine its own outputs, correcting incomplete reasoning steps and enhancing structural coherence.
This iterative refinement leads to more complete, logically consistent, and executable workflows.

\subsection{Discussion with Prior Works}
Across the growing body of research on ComfyUI~\cite{xu2025comfyuir1} workflow generation, there exists a common aspiration: to reduce the inherent complexity of workflow construction and to inject domain knowledge that enables structured reasoning.
Yet the ways the prior works operationalize this goal differ.

Early frameworks such as ComfyAgent~\cite{xue2025comfybench} and ComfyMind~\cite{guo2025comfymind} approach the challenge by introducing constrained abstractions.
They replaced raw node-level graphs with meta-nodes whose internal connectivity was predefined, lowering combinatorial complexity.
This approach, while controllable, inevitably limited expressiveness by constraining the solution space to handcrafted schemas.

Subsequent learning-based systems, including ComfyGPT~\cite{huang2025comfygpt} and ComfyUI-R1~\cite{xu2025comfyuir1}, shifted the focus from symbolic simplification to data-driven capability.
Instead of fixing structural templates, they trained language or reasoning models on large collections of workflows to predict connections or generate pseudo-codes directly from textual descriptions.
However, their knowledge remained implicitly entangled within model parameters, making it difficult to interpret, adapt, or reuse across levels of abstraction.

Our framework departs from both directions.
Rather than simplifying the problem or burying knowledge inside opaque model weights, we make knowledge itself the central representational currency.
Through knowledge inversion, we decompose complex workflows into hierarchical representations that capture structure, strategy, and reasoning.
This paradigm does not merely reduce complexity; it reorganizes it by transforming the implicit experiential knowledge of experts into explicit, learnable, and reusable forms.
In doing so, it moves workflow generation from ad-hoc pattern learning toward a principled, knowledge-centric foundation for agentic reasoning.

\section{Experiments}

\subsection{Settings}

\paragraph{Workflow Collection.}
We collected over 10,000 workflow instances from the Internet, covering a wide range of task types and node structures.
However, the raw data were highly noisy and redundant.
To construct a high-quality, diverse, and semantically consistent dataset, we performed large-scale structural deduplication and filtering.
Specifically, we first identified potentially redundant workflows by comparing node-level structural differences.
When the ratio of differing nodes between two workflows was below 10\%, they were considered structurally similar.
In such cases, we employed GPT-4o to evaluate semantic and procedural equivalence, retaining only the version with more standardized and widely used node types.
We then merged fully identical workflows by unifying their input entries and removed workflows that were either too simple (fewer than six nodes) or excessively complex (more than ninety nodes).
Task descriptions were further normalized and consolidated to eliminate near-duplicate objectives while maintaining structural and semantic diversity.

After multi-stage cleaning and filtering, we obtained a curated dataset of 912 high-quality workflows spanning diverse task distributions and node types. 
Among them, 882 workflows were used for training, while 30 workflows and their corresponding task queries were reserved for testing. 
The test workflows were grouped into three categories based on their similarity to the training data: \textit{Challenge 1} (high similarity), \textit{Challenge 2} (medium similarity), and \textit{Real-world Cases} (no comparable training samples). 
Similarity was assessed by a GPT-based evaluator across four dimensions: task objective, technical method, structural form, and application context.

In addition to this split, we further evaluate our method on two external benchmarks to assess generalization ability. 
First, we evaluate on the FlowBench dataset~\cite{huang2025comfygpt} using a test set of 30 tasks. 
Second, we conduct evaluation on a subset of ComfyBench consisting of 20 tasks sampled from the \textit{Creative} and \textit{Complex} categories. 
Together, these evaluations cover both in-domain and out-of-domain workflow generation scenarios.

\noindent\textbf{Compared Methods.} 
%%% Lei 讨论baseline的合理性
Given that ComfyBench \cite{xue2025comfybench} is the only open-source baseline currently available for this task, we adopt it for comparison and strictly follow its experimental settings. Specifically, we evaluate six representative workflow-generation paradigms established in \cite{xue2025comfybench}: Zero-Shot, Few-Shot, Chain-of-Thought (CoT), CoT with Self-Consistency (CoT-SC), Retrieval-Augmented Generation (RAG), and ComfyAgent. These methods correspond to the variants presented in Tables~\ref{tab:structure}, \ref{tab:pass_solve}, \ref{tab:comfybench_overall}, and \ref{tab:flow_bench_overall_results}.
% For the baselines, we adopt the workflow-generation paradigms proposed in~\cite{xue2025comfybench} and their variants as comparison methods, consistent with the settings used in the main paper.
% To ensure fairness, all methods use GPT-5 as the underlying LLM interface, consistent with our refine agent configuration.
% Following~\cite{xue2025comfybench}, we evaluate the following representative paradigms:
To ensure fairness, all the methods use GPT-5 as the underlying LLM interface, consistent with the configuration of our refine agent. 
More details of these baseline methods can be found in the supplementary material.

\noindent\textbf{Implementation detail.}
We used the Qwen3-14B model as the backbone for SFT training. The model was trained on the two tasks described above.
Both stages were fine-tuned using the LoRA technique on a single NVIDIA H200 GPU.
During the instruction-to-strategy stage, we set the maximum sequence length to 2048 tokens, while in the strategy-to-pseudocode stage, it was extended to 4096 tokens.
The LoRA configuration used a rank of 16 and a scaling factor ($\alpha$) of 32.
For the pre-trained large language model used for agent building, we adopt GPT-5 (gpt-5-turbo, 2025 release).

% \paragraph{ComfyUI Environment}

% \subsection{Baseline}
% %
% We adopt the methods proposed in~\cite{xue2025comfybench} and its related variants as baselines, as summarized in Table~\ref{tab:structure} and Table~\ref{tab:pass_solve}.
% %
% To ensure fairness, all comparative methods use GPT-5 as the underlying LLM interface, consistent with our refine agent configuration.
% %
% Following~\cite{xue2025comfybench}, we evaluate the following representative paradigms:

% \begin{itemize}
% \item \textbf{Zero-shot:} Directly prompts the LLM to generate a workflow from the task instruction alone.
% \item \textbf{Few-shot:} Extends the zero-shot setting by providing several in-context exemplars, allowing the model to better understand expected formats and improve generation accuracy~\cite{brown2020language}.
% \item \textbf{Chain-of-Thought (CoT):} Encourages the model to perform intermediate reasoning before producing the final workflow, leading to more structured and interpretable outputs~\cite{wei2022chain}.
% \item \textbf{CoT with Self-Consistency (CoT-SC):} Enhances CoT by sampling multiple reasoning paths and selecting the most self-consistent output as the final prediction~\cite{wang2022self}.
% \item \textbf{Retrieval-Augmented Generation (RAG):} Retrieves the most relevant exemplars from an external workflow corpus and conditions the LLM on these examples to ground its generation in prior knowledge~\cite{lewis2020retrieval}.
% \end{itemize}

\begin{figure*}[!t]
    \centering
    \includegraphics[width=0.93\linewidth]
    {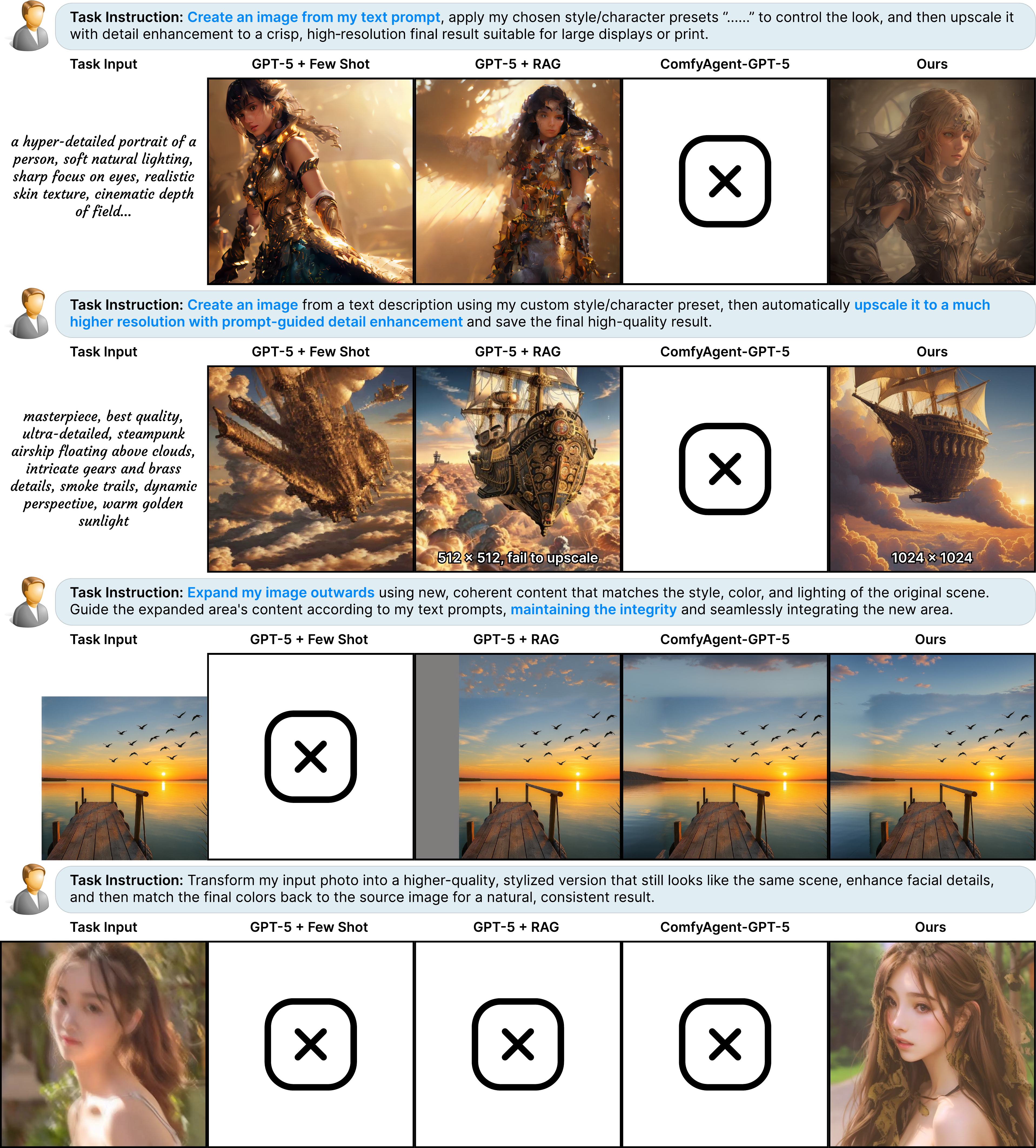}
    \caption{Comparison with baseline methods. Given a user instruction, our method accurately adheres to the task description.}
    \label{fig:vis-comp}
\end{figure*}

\subsection{Evaluation Metrics}
We comprehensively evaluate generated workflows from both \textit{structural} and \textit{semantic} perspectives using four complementary metrics:
Valid Node Diversity (VND), Node Completeness (NodeComp), Link Completeness (LinkComp), and Task Consistency (TaskCons).

\noindent \textbf{VND} quantifies the diversity of valid node types present in generated workflows, reflecting the model’s ability to produce varied and functionally valid module compositions rather than relying on repetitive or trivial components.

\noindent \textbf{NodeComp} measures the proportion of valid nodes that are correctly instantiated and involved in at least one connection. A node is considered valid if its type is registered, non-empty, and linked to other nodes. The ratio of valid to total nodes indicates the semantic and functional completeness of the workflow.

\noindent \textbf{LinkComp} evaluates the structural correctness of node connections. For each node, input–output consistency is verified: all declared inputs must exist and correctly reference valid outputs, and all output slot indices must remain within bounds. The proportion of nodes passing these checks represents the workflow’s topological soundness.

\noindent \textbf{TaskCons} assesses how well the generated workflow fulfills the target task described in the instruction. A GPT-based evaluator scores how well each workflow fulfills its task (e.g., ``upscale'', ``relighting'', ``pose-consistent animation''). Higher scores indicate better alignment between the task intent and the workflow’s functional structure.

In addition, we evaluate executability and task fulfillment through two outcome-level indicators: \textbf{Pass} and \textbf{Resolve}.
A workflow is counted as \textit{Pass} if it executes successfully in ComfyUI without structural or runtime errors, and as \textit{Resolve} if it additionally produces an output that fulfills the intended task goal. Together, these indicators assess both syntactic validity and the functional effectiveness of workflow generation.  To ensure stable and unbiased evaluation, we do not rely on LLM-based judges. Instead, all workflows are executed manually in ComfyUI and the outputs are assessed through human inspection.

\subsection{Results}
\textbf{Quantitative evaluation.} 
We evaluate workflow-generation quality on our test set using four complementary metrics: VND, NodeComp, LinkComp, and TaskCons. Together, these metrics measure workflow diversity, structural completeness, and task-level correctness. As shown in Table~\ref{tab:structure}, our method achieves clear improvements across all metrics. Our VND score reaches 152, more than 3.6 times the highest baseline score of 42, indicating that our system can compose substantially richer and more diverse workflow structures than template-based methods.
\begin{table}[!ht]
\centering
\caption{Comparison of workflow generation performance on the test set.
VND: Valid Node Diversity.
NodeComp: Node Completeness.
LinkComp: Link Completeness.
TaskCons: Task Consistency.}
\label{tab:structure}
\begin{tabular}{lcccc}
\toprule
Method & VND$\uparrow$ & NodeComp (\%)$\uparrow$ & LinkComp (\%)$\uparrow$ & TaskCons (\%)$\uparrow$ \\
\midrule
GPT-5 + Zero-Shot  & 0.0  & 0.0  & 0.0  & 0.0  \\
GPT-5 + Few-Shot   & 20   & 36.7 & 36.7 & 34.7 \\
GPT-5 + CoT        & 26   & 33.3 & 33.3 & 32.9 \\
GPT-5 + CoT-SC     & 21   & 36.7 & 36.7 & 34.8 \\
GPT-5 + RAG        & 42   & 56.7 & 56.7 & 53.6 \\
GPT-5 + ComfyAgent & 40   & 56.7 & 56.7 & 55.1 \\
\midrule
\textbf{Ours} & \textbf{152} & \textbf{96.3} & \textbf{98.3} & \textbf{86.9} \\
\bottomrule
\end{tabular}
\end{table}

Our method also attains markedly higher structural and semantic accuracy than all the other methods. Its NodeComp, LinkComp, and TaskCons scores are consistently far above GPT-5-based methods and ComfyAgent, demonstrating that the generated workflows are both well-formed and aligned with the intended task logic. These results show that template-based or retrieval-driven methods are fundamentally limited, while our model can generate diverse, structurally sound, and task-faithful workflows at scale.

As reported in Table~\ref{tab:pass_solve}, we further evaluate workflow executability within ComfyUI on our test set. Our approach achieves the highest pass and solve rates across all test categories—Challenge~1, Challenge~2, and Real-World Cases. For example, the average Pass rate reaches 86.9\%, compared with 36.4\% for ComfyAgent, and Resolve improves from 29.4\% to 61.1\%. 
These results indicate that our workflows are not only theoretically valid but also executable in practice and able to fulfill the specified task instruction. The strong performance on real-world cases further highlights the robustness of the generated pipelines.

\begin{table*}[!t]
\centering
\caption{Results grouped by test set categories.
\emph{Pass} indicates the generated workflow executes successfully without structural or runtime errors.
\emph{Resolve} indicates the workflow executes and fulfills the given instruction.}
\label{tab:pass_solve}
\resizebox{\textwidth}{!}{
\begin{tabular}{lcccccccc}
\toprule
\textbf{Method} &
\multicolumn{2}{c}{\textbf{Challenge 1}} &
\multicolumn{2}{c}{\textbf{Challenge 2}} &
\multicolumn{2}{c}{\textbf{Real-World Cases}} &
\multicolumn{2}{c}{\textbf{Average}} \\
\cmidrule(lr){2-3}\cmidrule(lr){4-5}\cmidrule(lr){6-7}\cmidrule(lr){8-9}
& Pass (\%) & Resolve (\%) & Pass (\%) & Resolve (\%) & Pass (\%) & Resolve (\%) & Pass (\%) & Resolve (\%) \\
\midrule
GPT-5 + Zero-Shot  & 0.0  & 0.0  & 0.0  & 0.0  & 0.0  & 0.0  & 0.0  & 0.0  \\
GPT-5 + Few-Shot   & 41.6 & 25.0 & 37.5 & 25.0 & 0.0  & 0.0  & 26.4 & 16.7 \\
GPT-5 + CoT        & 33.3 & 25.0 & 25.0 & 12.5 & 10.0 & 10.0 & 22.8 & 15.8 \\
GPT-5 + CoT-SC     & 25.0 & 25.0 & 37.5 & 12.5 & 10.0 & 0.0  & 24.2 & 12.5 \\
GPT-5 + RAG        & 66.7 & 41.7 & 37.5 & 25.0 & 40.0 & 10.0 & 48.1 & 25.6 \\
GPT-5 + ComfyAgent & 41.7 & 33.3 & 37.5 & 25.0 & 30.0 & 30.0 & 36.4 & 29.4 \\
\midrule
\textbf{Ours}      & \textbf{83.3} & \textbf{58.3} & \textbf{87.5} & \textbf{75.0} & \textbf{90.0} & \textbf{50.0} & \textbf{86.9} & \textbf{61.1} \\
\bottomrule
\end{tabular}
}
\end{table*}

We evaluate workflow-generation quality on ComfyBench~\cite{xue2025comfybench}, a benchmark of real-world visual-creation workflows that enables assessing model generalization. 
The evaluation is conducted on a held-out set of 20 tasks randomly sampled from the \emph{Creative} and \emph{Complex} categories. 
As shown in Table~\ref{tab:comfybench_overall}, our method substantially outperforms all the other methods across both structural and execution metrics. 
In particular, our model achieves a VND score of 138, which is over $3\times$ higher than the strongest baseline (43–45), indicating significantly richer workflow compositions. 
Structural correctness also improves markedly, with NodeComp increasing from 65.0\% to 79.8\% and LinkComp from 65.0\% to 82.7\%, while TaskCons rises from 59.3\% to 85.0\%. 
These structural gains translate into stronger execution reliability, yielding 50.0\% Pass and 25.0\% Resolve on average across the benchmark. 
Overall, these results show that our knowledge-centric framework generates workflows that are substantially more diverse, structurally coherent, and executable, demonstrating strong generalization on real-world workflow generation tasks.

\begin{table}[t]
\centering
\caption{Overall workflow generation performance on ComfyBench.
Workflow structure metrics include VND (Valid Node Diversity),
NodeComp (Node Completeness), LinkComp (Link Completeness),
and TaskCons (Task Consistency).
Execution metrics report the average Pass and Resolve rates.}
\label{tab:comfybench_overall}
\resizebox{\textwidth}{!}{
\begin{tabular}{lcccc|cc}
\toprule
& \multicolumn{4}{c|}{Workflow Structure} 
& \multicolumn{2}{c}{Execution} \\
Method
& VND$\uparrow$
& NodeComp (\%)$\uparrow$
& LinkComp (\%)$\uparrow$
& TaskCons (\%)$\uparrow$
& Pass (\%)$\uparrow$
& Resolve (\%)$\uparrow$ \\
\midrule
GPT-5 + Zero-Shot   & 0   & 0.0  & 0.0  & 0.0  & 0.0  & 0.0  \\
GPT-5 + Few-Shot    & 14  & 35.0 & 35.0 & 29.8 & 10.0 & 5.0  \\
GPT-5 + CoT         & 12  & 25.0 & 25.0 & 19.5 & 5.0  & 0.0  \\
GPT-5 + CoT-SC      & 18  & 20.0 & 20.0 & 17.1 & 0.0  & 0.0  \\
GPT-5 + RAG         & 45  & 60.0 & 60.0 & 57.7 & 40.0 & 25.0 \\
GPT-5 + ComfyAgent  & 43  & 65.0 & 65.0 & 59.3 & 35.0 & 20.0 \\
\midrule
Ours
& \textbf{138}
& \textbf{79.8}
& \textbf{82.7}
& \textbf{85.0}
& \textbf{50.0}
& \textbf{25.0} \\
\bottomrule
\end{tabular}
}
\end{table}

We evaluate our method on the FlowBench\cite{huang2025comfygpt} using a test set of 30 tasks, as summarized in Table~\ref{tab:flow_bench_overall_results}. 
Our approach consistently outperforms all all the other methods across both workflow structure and execution metrics. 
In particular, it achieves a VND score of 145, substantially higher than the strongest baseline (41–42), indicating significantly richer workflow compositions. 
Structural correctness is also markedly improved, with NodeComp reaching 94.4\% and LinkComp 98.0\%, compared to around 53–73\% for existing methods, while TaskCons improves to 93.4\%. 
These structural advantages translate into stronger execution reliability, achieving 66.7\% Pass and 36.7\% Solve rates, exceeding all baseline approaches. Overall, the results demonstrate that our method achieves strong generalization on the  FlowBench\cite{huang2025comfygpt} while consistently outperforming existing all the other methods.

% \section{LLM Judge Bias Analysis}
% \label{ask consistency evaluation}

% \begin{table}[t]
% \centering
% \caption{Overall performance comparison on workflow generation.
% Workflow Structure metrics include VND (Valid Node Diversity),
% NodeComp (Node Completeness), LinkComp (Link Completeness),
% and TaskCons (Task Consistency).
% Execution metrics include Pass and Solve rates.}
% \label{tab:overall_results}
% \resizebox{\textwidth}{!}{
% \begin{tabular}{lcccccc}
% \toprule
% Method
% & VND$\uparrow$
% & NodeComp (\%)$\uparrow$
% & LinkComp (\%)$\uparrow$
% & TaskCons (\%)$\uparrow$
% & Pass (\%)$\uparrow$
% & Solve (\%)$\uparrow$ \\
% \midrule
% Zero-Shot   & 0  & 0.00  & 0.00  & 0.00  & 0.0 & 0.0 \\
% Few-Shot    & 27 & 46.67 & 46.67 & 45.50 & 26.7 & 16.7 \\
% CoT         & 24 & 43.33 & 43.33 & 42.10 & 23.3 & 16.7 \\
% CoT-SC      & 26 & 40.00 & 40.00 & 38.20 & 20.0 & 13.3 \\
% RAG         & 42 & 73.33 & 73.33 & 67.43 & 46.7 & 26.7 \\
% ComfyGPT-5  & 41 & 53.33 & 53.33 & 50.80 & 40.0 & 26.7 \\
% \midrule
% \textbf{Ours}
% & \textbf{145}
% & \textbf{94.4}
% & \textbf{98.0}
% & \textbf{93.4}
% & \textbf{66.7}
% & \textbf{36.7} \\
% \bottomrule
% \end{tabular}
% }
% \end{table}
\begin{table}[t]

\centering
\caption{Overall performance comparison on FlowBench\cite{huang2025comfygpt}.
Read as Tab.\ref{tab:comfybench_overall}.}
\label{tab:flow_bench_overall_results}
\resizebox{\textwidth}{!}{
\begin{tabular}{lcccc|cc}
\toprule
& \multicolumn{4}{c|}{Workflow Structure}
& \multicolumn{2}{c}{Execution} \\
Method
& VND$\uparrow$
& NodeComp (\%)$\uparrow$
& LinkComp (\%)$\uparrow$
& TaskCons (\%)$\uparrow$
& Pass (\%)$\uparrow$
& Solve (\%)$\uparrow$ \\
\midrule
GPT-5 + Zero-Shot   & 0  & 0.00  & 0.00  & 0.00  & 0.0 & 0.0 \\
GPT-5 + Few-Shot    & 27 & 46.67 & 46.67 & 45.50 & 26.7 & 16.7 \\
GPT-5 + CoT         & 24 & 43.33 & 43.33 & 42.10 & 23.3 & 16.7 \\
GPT-5 + CoT-SC      & 26 & 40.00 & 40.00 & 38.20 & 20.0 & 13.3 \\
GPT-5 + RAG         & 42 & 73.33 & 73.33 & 67.43 & 46.7 & 26.7 \\
GPT-5 + ComfyAgent  & 41 & 53.33 & 53.33 & 50.80 & 40.0 & 26.7 \\
\midrule
Ours
& \textbf{145}
& \textbf{94.4}
& \textbf{98.0}
& \textbf{93.4}
& \textbf{66.7}
& \textbf{36.7} \\
\bottomrule
\end{tabular}
}
\end{table}

\noindent \textbf{LLM Judge Bias Analysis.} To examine the robustness of task consistency evaluation, we compare two LLM judges, GPT-5 and Gemini-2.5, on the same set of generated workflows. 
As shown in Table~\ref{tab:task_consistency}, both evaluators produce highly consistent performance trends across prompting strategies. 
Our method achieves the highest task consistency under both GPT-5 (86.9\%) and Gemini-2.5 (85.6\%), while the relative ranking of competing approaches remains largely unchanged. 
This consistency suggests that the evaluation results are not sensitive to the choice of LLM judge and are unlikely to be dominated by model-specific bias. 
In addition, execution-based metrics such as Pass and Solve provide an independent signal of task correctness, further supporting the reliability of the evaluation.

\begin{table}[ht]
\centering
\caption{Task consistency (\%) evaluated by two LLM judges (GPT-5 and Gemini-2.5) across different methods. Higher is better.}
\label{tab:task_consistency}

\small
\setlength{\tabcolsep}{4pt}

\begin{tabular}{lccccccc}
\toprule
Model
& Zero-Shot
& Few-Shot
& CoT
& CoT-SC
& RAG
& ComfyAgent
& Ours \\
\midrule
Gemini-2.5
& 0
& 34.0
& 30.8
& 34.2
& 52.0
& 40.8
& \textbf{85.6} \\

GPT-5
& 0
& 34.7
& 32.9
& 34.8
& 53.6
& 55.1
& \textbf{86.9} \\
\bottomrule
\end{tabular}
\end{table}

\noindent{\textbf{Effect of Hierarchical Workflow Modeling.}}
To provide a strong training-based baseline, we train Qwen3-32B end-to-end to directly map user instructions to pseudocode workflows (Q2P).  As shown in Table~\ref{tab:qwen_comparison}, our method with Qwen3-14B consistently outperforms the Qwen3-32B Q2P baseline across both workflow structure and execution metrics, despite using a smaller backbone model.  In particular, our approach improves VND from 148 to 152, NodeComp from 92.3\% to 96.3\%, and LinkComp from 95.8\% to 98.3\%, indicating more complete and structurally coherent workflow generation.  The advantage becomes even more pronounced in execution reliability: Pass increases from 56.7\% to 86.9\% and Solve from 33.3\% to 61.1\%, demonstrating that our method produces workflows that are substantially more executable and task-consistent. These results highlight the benefit of the proposed hierarchical design compared with direct end-to-end training.  Instead of learning a single mapping from instructions to workflows, our framework decomposes workflows into multi-level supervision signals, including high-level design strategies, intermediate reasoning traces, and structural skeleton topologies. 
This structured supervision enables the model to capture workflow design principles explicitly and to compose modules in a more systematic manner, leading to better structural fidelity and higher execution success rates.

\begin{table}[!t]

\centering
\caption{Comparison with Qwen3-32B Q2P on workflow generation.
Read as Tab. \ref{tab:comfybench_overall}}
\label{tab:qwen_comparison}

\resizebox{0.75\linewidth}{!}{
\begin{tabular}{lccccc}
\toprule
& \multicolumn{3}{c}{Workflow Structure} 
& \multicolumn{2}{c}{Execution} \\
\cmidrule(lr){2-4}\cmidrule(lr){5-6}
Method
& VND$\uparrow$
& NodeComp (\%)$\uparrow$
& LinkComp (\%)$\uparrow$
& Pass (\%)$\uparrow$
& Solve (\%)$\uparrow$ \\
\midrule
Qwen3-32B Q2P
& 148
& 92.3
& 95.8
& 56.7
& 33.3 \\
Ours
& \textbf{152}
& \textbf{96.3}
& \textbf{98.3}
& \textbf{86.9}
& \textbf{61.1} \\
\bottomrule
\end{tabular}
}
\end{table}

\noindent \textbf{Qualitative evaluation.}
As shown in Fig.~\ref{fig:vis-comp}, our method accurately interprets concise task instructions and assembles suitable node compositions across diverse tasks, including text-to-image generation, outpainting, face restoration, and style transfer. The qualitative results show that our agent generates more complete and better-structured workflows, producing higher-quality visual outputs than ComfyAgent and the other methods.

\noindent \textbf{Limitations and Discussion.} We observe a limitation when tasks require relatively rare node types. In such cases, the model tends to substitute them with more common nodes, which may break functional dependencies and increase structural errors in the generated workflow. We attribute this behavior mainly to limited coverage of rare nodes in the training data; expanding the diversity and scale of training workflows could help mitigate this issue. In addition, incorporating reinforcement-style optimization (e.g., GRPO) may further improve robustness.

\section{Conclusion}
In this work, we present a knowledge-centric framework for workflow generation in visual creation systems such as ComfyUI. Unlike prior LLM-based approaches that treat workflow synthesis as a direct text-to-JSON prediction problem, our method explicitly models the structure, hierarchy, and dynamics of expert knowledge.Through knowledge inversion, the model learns multilevel representations, including pseudo-codes, structural skeletons, and design strategies, from large collections of real workflows. Knowledge injection then guides the model to map tasks to strategies and strategies to executable structures via supervised reasoning. During inference, reversible reasoning and self-refinement enable the generation of coherent, executable workflows. Extensive experiments show that our approach produces more diverse nodes, more consistent compositions, and significantly higher execution success rates than existing methods. Overall, this work establishes a new direction for knowledge-driven, agentic workflow generation in visual creation systems.

\section*{Acknowledge}
This research was partially funded by the Ministry of Education and Science of Bulgaria, through support for INSAIT as part of the Bulgarian National Roadmap for Research Infrastructure. This project is also partially supported by Adobe Research.

\bibliographystyle{splncs04}
\bibliography{main}

@String(CVPR  = {IEEE Conf. Comput. Vis. Pattern Recog.})

@String(NeurIPS = {Adv. Neural Inform. Process. Syst.})

@String(AAAI  = {AAAI})

@String(CVPR  = {CVPR})

@String(NeurIPS = {NeurIPS})

@inproceedings{xue2025comfybench,
  title={Comfybench: Benchmarking llm-based agents in comfyui for autonomously designing collaborative ai systems},
  author={Xue, Xiangyuan and Lu, Zeyu and Huang, Di and Wang, Zidong and Ouyang, Wanli and Bai, Lei},
  booktitle={Proceedings of the Computer Vision and Pattern Recognition Conference},
  pages={24614--24624},
  year={2025}
}

@article{huang2025comfygpt,
  title={ComfyGPT: A Self-Optimizing Multi-Agent System for Comprehensive ComfyUI Workflow Generation},
  author={Huang, Oucheng and Ma, Yuhang and Zhao, Zeng and Wu, Mingrui and Ji, Jiayi and Zhang, Rongsheng and Hu, Zhipeng and Sun, Xiaoshuai and Ji, Rongrong},
  journal={arXiv preprint arXiv:2503.17671},
  year={2025}
}

@article{xu2025comfyuir1,
  title={ComfyUI-R1: Exploring Reasoning Models for Workflow Generation},
  author={Xu, Zhenran and Wang, Yiyu and Yang, Xue and Wang, Longyue and Luo, Weihua and Zhang, Kaifu and Hu, Baotian and Zhang, Min},
  journal={arXiv preprint arXiv:2506.09790},
  year={2025}
}

@inproceedings{chen2025symbolic,
  title={Symbolic Representation for Any-to-Any Generative Tasks},
  author={Chen, Jiaqi and Zhu, Xiaoye and Wang, Yue and Liu, Tianyang and Chen, Xinhui and Chen, Ying and Leong, Chak Tou and Ke, Yifei and Liu, Joseph and Yuan, Yiwen and others},
  booktitle={Proceedings of the Computer Vision and Pattern Recognition Conference},
  pages={27816--27826},
  year={2025}
}

@article{guo2025comfymind,
  title={ComfyMind: Toward General-Purpose Generation via Tree-Based Planning and Reactive Feedback},
  author={Guo, Litao and Xu, Xinli and Wang, Luozhou and Lin, Jiantao and Zhou, Jinsong and Zhang, Zixin and Su, Bolan and Chen, Ying-Cong},
  journal={arXiv preprint arXiv:2505.17908},
  year={2025}
}

@article{sobania2024comfygi,
  title={Comfygi: Automatic improvement of image generation workflows},
  author={Sobania, Dominik and Briesch, Martin and Rothlauf, Franz},
  journal={arXiv preprint arXiv:2411.14193},
  year={2024}
}

@misc{comfyanonymous2023comfyui,
  author       = {comfyanonymous},
  title        = {ComfyUI},
  year         = {2023},
  howpublished = {\url{https://github.com/comfyanonymous/ComfyUI}},
}

@inproceedings{NEURIPS2020_4c5bcfec_diffusion,
 author = {Ho, Jonathan and Jain, Ajay and Abbeel, Pieter},
 booktitle = {Advances in Neural Information Processing Systems},
 editor = {H. Larochelle and M. Ranzato and R. Hadsell and M.F. Balcan and H. Lin},
 pages = {6840--6851},
 publisher = {Curran Associates, Inc.},
 title = {Denoising Diffusion Probabilistic Models},
 url = {https://proceedings.neurips.cc/paper_files/paper/2020/file/4c5bcfec8584af0d967f1ab10179ca4b-Paper.pdf},
 volume = {33},
 year = {2020}
}

@inproceedings{NEURIPS2021_49ad23d1_diffusion,
 author = {Dhariwal, Prafulla and Nichol, Alexander},
 booktitle = {Advances in Neural Information Processing Systems},
 editor = {M. Ranzato and A. Beygelzimer and Y. Dauphin and P.S. Liang and J. Wortman Vaughan},
 pages = {8780--8794},
 publisher = {Curran Associates, Inc.},
 title = {Diffusion Models Beat GANs on Image Synthesis},
 url = {https://proceedings.neurips.cc/paper_files/paper/2021/file/49ad23d1ec9fa4bd8d77d02681df5cfa-Paper.pdf},
 volume = {34},
 year = {2021}
}

@inproceedings{kumari2023multi,
  title={Multi-concept customization of text-to-image diffusion},
  author={Kumari, Nupur and Zhang, Bingliang and Zhang, Richard and Shechtman, Eli and Zhu, Jun-Yan},
  booktitle={Proceedings of the IEEE/CVF Conference on Computer Vision and Pattern Recognition},
  pages={1931--1941},
  year={2023}
}

@article{gal2024comfygen,
  title={Comfygen: Prompt-adaptive workflows for text-to-image generation},
  author={Gal, Rinon and Haviv, Adi and Alaluf, Yuval and Bermano, Amit H and Cohen-Or, Daniel and Chechik, Gal},
  journal={arXiv preprint arXiv:2410.01731},
  year={2024}
}

@inproceedings{ruiz2023dreambooth,
  title={Dreambooth: Fine tuning text-to-image diffusion models for subject-driven generation},
  author={Ruiz, Nataniel and Li, Yuanzhen and Jampani, Varun and Pritch, Yael and Rubinstein, Michael and Aberman, Kfir},
  booktitle={Proceedings of the IEEE/CVF Conference on Computer Vision and Pattern Recognition},
  pages={22500--22510},
  year={2023}
}

@article{li2023dreamedit,
  title={DreamEdit: Subject-driven Image Editing},
  author={Li, Tianle and Ku, Max and Wei, Cong and Chen, Wenhu},
  journal={arXiv preprint arXiv:2306.12624},
  year={2023}
}

@inproceedings{zhang2023adding,
  title={Adding conditional control to text-to-image diffusion models},
  author={Zhang, Lvmin and Rao, Anyi and Agrawala, Maneesh},
  booktitle={CVPR},
  year={2023}
}

@article{xu2025comfyuicopilot,
  title={ComfyUI-Copilot: An Intelligent Assistant for Automated Workflow Development},
  author={Xu, Zhenran and Yang, Xue and Wang, Yiyu and Hu, Qingli and Wu, Zijiao and Wang, Longyue and Luo, Weihua and Zhang, Kaifu and Hu, Baotian and Zhang, Min},
  journal={arXiv preprint arXiv:2506.05010},
  year={2025}
}

@article{brown2020language,
  title={Language models are few-shot learners},
  author={Brown, Tom and Mann, Benjamin and Ryder, Nick and Subbiah, Melanie and Kaplan, Jared D and Dhariwal, Prafulla and Neelakantan, Arvind and Shyam, Pranav and Sastry, Girish and Askell, Amanda and others},
  journal={Advances in neural information processing systems},
  volume={33},
  pages={1877--1901},
  year={2020}
}

@inproceedings{devlin2019bert,
  title={Bert: Pre-training of deep bidirectional transformers for language understanding},
  author={Devlin, Jacob and Chang, Ming-Wei and Lee, Kenton and Toutanova, Kristina},
  booktitle={Proceedings of the 2019 conference of the North American chapter of the association for computational linguistics: human language technologies, volume 1 (long and short papers)},
  pages={4171--4186},
  year={2019}
}

@misc{qian2025toolrl,
      title={ToolRL: Reward is All Tool Learning Needs}, 
      author={Cheng Qian and Emre Can Acikgoz and Qi He and Hongru Wang and Xiusi Chen and Dilek Hakkani-Tür and Gokhan Tur and Heng Ji},
      year={2025},
      eprint={2504.13958},
      archivePrefix={arXiv},
      primaryClass={cs.LG},
      url={https://arxiv.org/abs/2504.13958}, 
}

@misc{li2025torlscalingtoolintegratedrl,
      title={ToRL: Scaling Tool-Integrated RL}, 
      author={Xuefeng Li and Haoyang Zou and Pengfei Liu},
      year={2025},
      eprint={2503.23383},
      archivePrefix={arXiv},
      primaryClass={cs.CL},
      url={https://arxiv.org/abs/2503.23383}, 
}

@article{wei2022chain,
  title={Chain-of-thought prompting elicits reasoning in large language models},
  author={Wei, Jason and Wang, Xuezhi and Schuurmans, Dale and Bosma, Maarten and Xia, Fei and Chi, Ed and Le, Quoc V and Zhou, Denny and others},
  journal={Advances in neural information processing systems},
  volume={35},
  pages={24824--24837},
  year={2022}
}

@article{wang2022self,
  title={Self-consistency improves chain of thought reasoning in language models},
  author={Wang, Xuezhi and Wei, Jason and Schuurmans, Dale and Le, Quoc and Chi, Ed and Narang, Sharan and Chowdhery, Aakanksha and Zhou, Denny},
  journal={arXiv preprint arXiv:2203.11171},
  year={2022}
}

@inproceedings{franklin1996agent,
  title={Is it an Agent, or just a Program?: A Taxonomy for Autonomous Agents},
  author={Franklin, Stan and Graesser, Art},
  booktitle={International workshop on agent theories, architectures, and languages},
  pages={21--35},
  year={1996},
  organization={Springer}
}

@article{li2024multimodal,
  title={Multimodal foundation models: From specialists to general-purpose assistants},
  author={Li, Chunyuan and Gan, Zhe and Yang, Zhengyuan and Yang, Jianwei and Li, Linjie and Wang, Lijuan and Gao, Jianfeng and others},
  journal={Foundations and Trends{\textregistered} in Computer Graphics and Vision},
  volume={16},
  number={1-2},
  pages={1--214},
  year={2024},
  publisher={Now Publishers, Inc.}
}

@article{silver2018general,
  title={A general reinforcement learning algorithm that masters chess, shogi, and Go through self-play},
  author={Silver, David and Hubert, Thomas and Schrittwieser, Julian and Antonoglou, Ioannis and Lai, Matthew and Guez, Arthur and Lanctot, Marc and Sifre, Laurent and Kumaran, Dharshan and Graepel, Thore and others},
  journal={Science},
  volume={362},
  number={6419},
  pages={1140--1144},
  year={2018},
  publisher={American Association for the Advancement of Science}
}

@article{hwangbo2019learning,
  title={Learning agile and dynamic motor skills for legged robots},
  author={Hwangbo, Jemin and Lee, Joonho and Dosovitskiy, Alexey and Bellicoso, Dario and Tsounis, Vassilios and Koltun, Vladlen and Hutter, Marco},
  journal={Science Robotics},
  volume={4},
  number={26},
  pages={eaau5872},
  year={2019},
  publisher={American Association for the Advancement of Science}
}

@article{chen2023autoagents,
  title={Autoagents: A framework for automatic agent generation},
  author={Chen, Guangyao and Dong, Siwei and Shu, Yu and Zhang, Ge and Sesay, Jaward and Karlsson, B{\"o}rje F and Fu, Jie and Shi, Yemin},
  journal={arXiv preprint arXiv:2309.17288},
  year={2023}
}

@article{achiam2023gpt,
  title={Gpt-4 technical report},
  author={Achiam, Josh and Adler, Steven and Agarwal, Sandhini and Ahmad, Lama and Akkaya, Ilge and Aleman, Florencia Leoni and Almeida, Diogo and Altenschmidt, Janko and Altman, Sam and Anadkat, Shyamal and others},
  journal={arXiv preprint arXiv:2303.08774},
  year={2023}
}

@article{yang2025qwen3,
  title={Qwen3 technical report},
  author={Yang, An and Li, Anfeng and Yang, Baosong and Zhang, Beichen and Hui, Binyuan and Zheng, Bo and Yu, Bowen and Gao, Chang and Huang, Chengen and Lv, Chenxu and others},
  journal={arXiv preprint arXiv:2505.09388},
  year={2025}
}

@misc{cursor2024,
  title        = {Cursor: AI-powered Code Editor},
  author={Cursor},
  howpublished = {\url{https://www.cursor.com/en}},
}

@article{comanici2025gemini,
  title={Gemini 2.5: Pushing the frontier with advanced reasoning, multimodality, long context, and next generation agentic capabilities},
  author={Comanici, Gheorghe and Bieber, Eric and Schaekermann, Mike and Pasupat, Ice and Sachdeva, Noveen and Dhillon, Inderjit and Blistein, Marcel and Ram, Ori and Zhang, Dan and Rosen, Evan and others},
  journal={arXiv preprint arXiv:2507.06261},
  year={2025}
}

@article{touvron2023llama,
  title={Llama: Open and efficient foundation language models},
  author={Touvron, Hugo and Lavril, Thibaut and Izacard, Gautier and Martinet, Xavier and Lachaux, Marie-Anne and Lacroix, Timoth{\'e}e and Rozi{\`e}re, Baptiste and Goyal, Naman and Hambro, Eric and Azhar, Faisal and others},
  journal={arXiv preprint arXiv:2302.13971},
  year={2023}
}

@article{team2023gemini,
  title={Gemini: a family of highly capable multimodal models},
  author={Team, Gemini and Anil, Rohan and Borgeaud, Sebastian and Alayrac, Jean-Baptiste and Yu, Jiahui and Soricut, Radu and Schalkwyk, Johan and Dai, Andrew M and Hauth, Anja and Millican, Katie and others},
  journal={arXiv preprint arXiv:2312.11805},
  year={2023}
}

@article{luo2022biogpt,
  title={BioGPT: generative pre-trained transformer for biomedical text generation and mining},
  author={Luo, Renqian and Sun, Liai and Xia, Yingce and Qin, Tao and Zhang, Sheng and Poon, Hoifung and Liu, Tie-Yan},
  journal={Briefings in bioinformatics},
  volume={23},
  number={6},
  pages={bbac409},
  year={2022},
  publisher={Oxford University Press}
}

@article{singhal2023large,
  title={Large language models encode clinical knowledge},
  author={Singhal, Karan and Azizi, Shekoofeh and Tu, Tao and Mahdavi, S Sara and Wei, Jason and Chung, Hyung Won and Scales, Nathan and Tanwani, Ajay and Cole-Lewis, Heather and Pfohl, Stephen and others},
  journal={Nature},
  volume={620},
  number={7972},
  pages={172--180},
  year={2023},
  publisher={Nature Publishing Group}
}

@article{taylor2022galactica,
  title={Galactica: A large language model for science},
  author={Taylor, Ross and Kardas, Marcin and Cucurull, Guillem and Scialom, Thomas and Hartshorn, Anthony and Saravia, Elvis and Poulton, Andrew and Kerkez, Viktor and Stojnic, Robert},
  journal={arXiv preprint arXiv:2211.09085},
  year={2022}
}

@misc{shao2024deepseekmath,
      title={DeepSeekMath: Pushing the Limits of Mathematical Reasoning in Open Language Models}, 
      author={Zhihong Shao and Peiyi Wang and Qihao Zhu and Runxin Xu and Junxiao Song and Xiao Bi and Haowei Zhang and Mingchuan Zhang and Y. K. Li and Y. Wu and Daya Guo},
      year={2024},
      eprint={2402.03300},
      archivePrefix={arXiv},
      primaryClass={cs.CL},
      url={https://arxiv.org/abs/2402.03300}, 
}

@article{yao2024tree,
  title={Tree of thoughts: Deliberate problem solving with large language models},
  author={Yao, Shunyu and Yu, Dian and Zhao, Jeffrey and Shafran, Izhak and Griffiths, Tom and Cao, Yuan and Narasimhan, Karthik},
  journal={Advances in Neural Information Processing Systems},
  volume={36},
  year={2024}
}

@inproceedings{besta2024graph,
  title={Graph of thoughts: Solving elaborate problems with large language models},
  author={Besta, Maciej and Blach, Nils and Kubicek, Ales and Gerstenberger, Robert and Podstawski, Michal and Gianinazzi, Lukas and Gajda, Joanna and Lehmann, Tomasz and Niewiadomski, Hubert and Nyczyk, Piotr and others},
  booktitle={Proceedings of the AAAI Conference on Artificial Intelligence},
  volume={38},
  number={16},
  pages={17682--17690},
  year={2024}
}

@article{shen2024hugginggpt,
  title={Hugginggpt: Solving ai tasks with chatgpt and its friends in hugging face},
  author={Shen, Yongliang and Song, Kaitao and Tan, Xu and Li, Dongsheng and Lu, Weiming and Zhuang, Yueting},
  journal={Advances in Neural Information Processing Systems},
  volume={36},
  year={2024}
}

@article{wu2023visual,
  title={Visual chatgpt: Talking, drawing and editing with visual foundation models},
  author={Wu, Chenfei and Yin, Shengming and Qi, Weizhen and Wang, Xiaodong and Tang, Zecheng and Duan, Nan},
  journal={arXiv preprint arXiv:2303.04671},
  year={2023}
}

@article{shinn2024reflexion,
  title={Reflexion: Language agents with verbal reinforcement learning},
  author={Shinn, Noah and Cassano, Federico and Gopinath, Ashwin and Narasimhan, Karthik and Yao, Shunyu},
  journal={Advances in Neural Information Processing Systems},
  volume={36},
  year={2024}
}

@inproceedings{lewis2020retrieval,
  title={Retrieval-augmented generation for knowledge-intensive nlp tasks},
  author={Lewis, Patrick and Perez, Ethan and Piktus, Aleksandra and Petroni, Fabio and Karpukhin, Vladimir and Goyal, Naman and K{\"u}ttler, Heinrich and Lewis, Mike and Yih, Wen-tau and Rockt{\"a}schel, Tim and others},
  booktitle={NeurIPS},
  year={2020}
}

@inproceedings{zhuge2024gptswarm,
  title={Gptswarm: Language agents as optimizable graphs},
  author={Zhuge, Mingchen and Wang, Wenyi and Kirsch, Louis and Faccio, Francesco and Khizbullin, Dmitrii and Schmidhuber, J{\"u}rgen},
  booktitle={Forty-first International Conference on Machine Learning},
  year={2024}
}

@article{yuksekgonul2024textgrad,
  title={Textgrad: Automatic" differentiation" via text},
  author={Yuksekgonul, Mert and Bianchi, Federico and Boen, Joseph and Liu, Sheng and Huang, Zhi and Guestrin, Carlos and Zou, James},
  journal={arXiv preprint arXiv:2406.07496},
  year={2024}
}

@article{zhuge2024agent,
  title={Agent-as-a-judge: Evaluate agents with agents},
  author={Zhuge, Mingchen and Zhao, Changsheng and Ashley, Dylan and Wang, Wenyi and Khizbullin, Dmitrii and Xiong, Yunyang and Liu, Zechun and Chang, Ernie and Krishnamoorthi, Raghuraman and Tian, Yuandong and others},
  journal={arXiv preprint arXiv:2410.10934},
  year={2024}
}

@inproceedings{chen2024humans,
  title={Humans or LLMs as the judge? a study on judgement bias},
  author={Chen, Guiming Hardy and Chen, Shunian and Liu, Ziche and Jiang, Feng and Wang, Benyou},
  booktitle={Proceedings of the 2024 Conference on Empirical Methods in Natural Language Processing},
  pages={8301--8327},
  year={2024}
}

\clearpage

\begin{center}
{\LARGE\bfseries Knowledge-Centric Agents for Workflow Generation in ComfyUI}\\[0.5em]
{\Large\itshape Supplementary Materials}
\end{center}

\vspace{1em}

\appendix

This supplementary material is organized as follows.
% First, \S\ref{ask consistency evaluation} analyzes potential bias in GPT-based judges for task consistency evaluation.
First, \S\ref{workflow ana} presents qualitative workflow structure comparisons through representative case studies.
Second, \S\ref{sec:comparison methods} summarizes the comparison methods used in our experiments, including zero-shot prompting, few-shot prompting, reasoning-based strategies, and retrieval-augmented methods.
Finally, \S\ref{sec:dataset_supp} provides a detailed analysis of task distributions and workflow compositions in the training and test datasets.

\section{Workflow Structure Analysis}
 \label{workflow ana}
\begin{figure}[!t]
\centering
\includegraphics[width=0.8\linewidth]{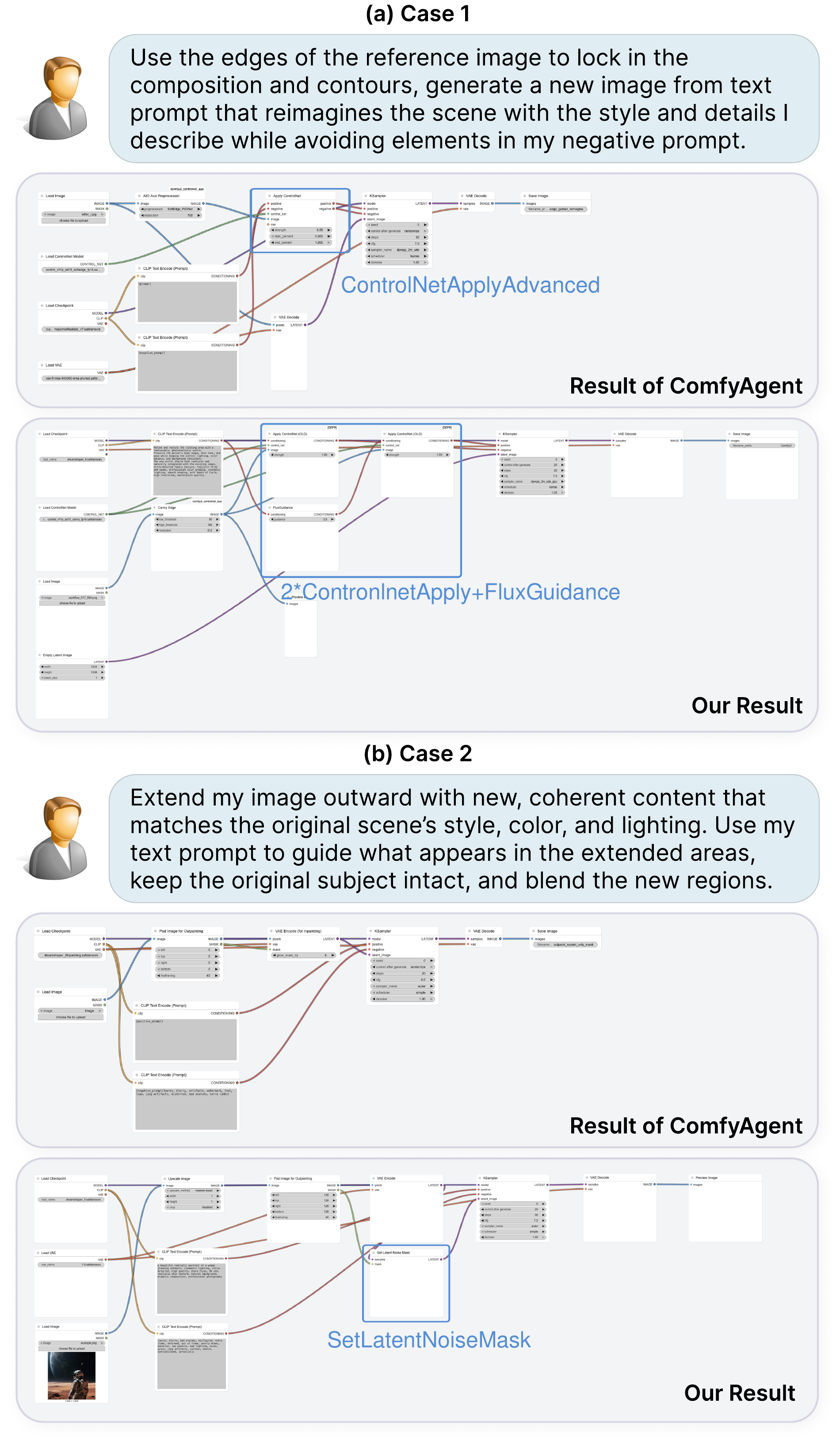}
\caption{Comparison of generated workflow structures.}
\label{fig:cases}
\end{figure}

As illustrated in Fig.\ref{fig:cases}, for \textbf{Case 1}, the workflow generated by our method adopts a design that uses \emph{two} \texttt{ControlNetApply} modules together with a \texttt{FluxGuidance} module, while ComfyAgent's workflow is based on a single \texttt{ControlNetApply\-Advanced}.  
This design offers several advantages: employing two ControlNet modules allows the workflow to simultaneously preserve structural fidelity and maintain smoother stylistic contours—two objectives that are difficult to balance using a single ControlNet, since it only provides one strength parameter.  
In addition, the \texttt{FluxGuidance} component further enhances textual alignment and leads to more consistent stylistic control. For \textbf{Case 2}, the workflow produced by our method incorporates ~\texttt{SetLatentNoiseMask}, which assigns noise to the latent representation of the out painting region. This enables the model to freely generate new content in the extended areas while keeping the original regions intact.

Taken together, these two cases reveal that, under the same task instruction, ComfyAgent typically falls back to a preset, simplified template that often deviates from real task requirements. In contrast, workflows generated by our method show greater diversity and stronger task-specific reasoning. This stems from learning rich workflow-design knowledge from large-scale data, enabling the model to understand effective functional compositions—such as when to use paired ControlNet modules—and make more intelligent design decisions.

\section{Comparison Methods}
\label{sec:comparison methods}
We adopt the workflow-generation paradigms proposed in~\cite{xue2025comfybench} and their variants as comparison methods, consistent with the settings used in the main paper.
To ensure fairness, all methods use GPT-5 as the underlying LLM interface, consistent with our refine agent configuration.
Following~\cite{xue2025comfybench}, we evaluate the following representative paradigms:

\begin{itemize}
\item \textbf{Zero-shot:} Directly prompts the LLM to generate a workflow from the task instruction alone.
\item \textbf{Few-shot:} Extends the zero-shot setting by providing several in-context exemplars, allowing the model to better understand expected formats and improve generation accuracy~\cite{brown2020language}.
\item \textbf{Chain-of-Thought (CoT):} Encourages the model to perform intermediate reasoning before producing the final workflow, leading to more structured and interpretable outputs~\cite{wei2022chain}.
\item \textbf{CoT with Self-Consistency (CoT-SC):} Enhances CoT by sampling multiple reasoning paths and selecting the most self-consistent output as the final prediction~\cite{wang2022self}.
\item \textbf{Retrieval-Augmented Generation (RAG):} Retrieves the most relevant exemplars from an external workflow corpus and conditions the LLM on these examples to ground its generation in prior knowledge~\cite{lewis2020retrieval}.
\end{itemize}

\section{Dataset Analysis}
\label{sec:dataset_supp}
To comprehensively characterize the task distribution and workflow complexity in our dataset, we conducted a systematic analysis of task types across both the training and test datasets. Our analysis employed GPT-5 for multi-label classification, recognizing that a single workflow may encompass multiple task types (e.g., a workflow that generates an image and then upscales it would be classified as both \texttt{text\_to\_image} and \texttt{upscaling}).

\subsection{Training Dataset Analysis}

\begin{figure*}[!tbp]
    \centering
    \includegraphics[width=1\textwidth]{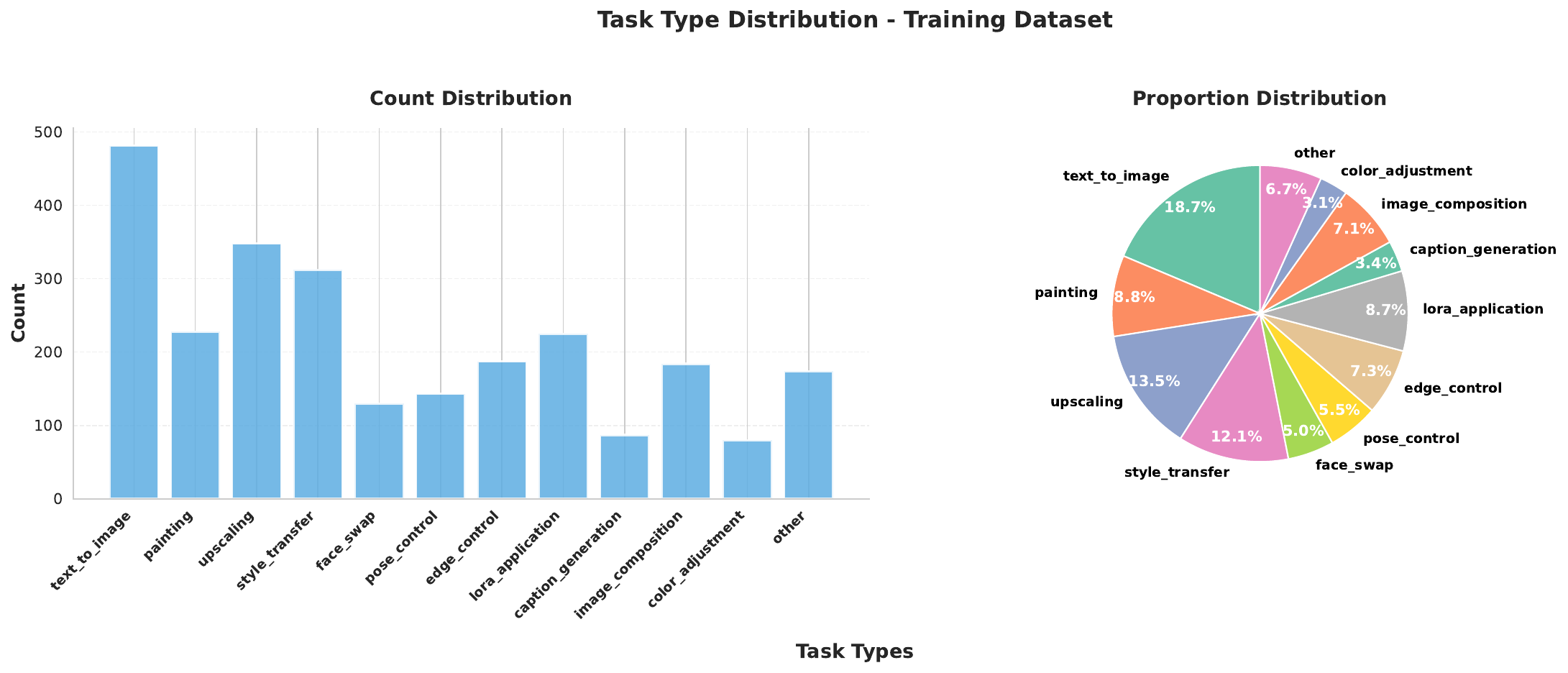}
\caption{Task type distribution in the training dataset. Left: count of workflows containing each task type (workflow-level distribution, showing the number of workflows with each task type). Right: proportion of each task type among all task type occurrences (occurrence-level distribution, accounting for multi-label classification where each workflow may contain multiple task types).}
    \label{fig:task_combined_train}
\end{figure*}

The training dataset comprises 879 workflow items with 635 unique node types, demonstrating the rich variety of components utilized in image generation pipelines. Task type classification reveals that the majority of workflows involve composite operations combining multiple task types, with an average of 2.93 task types per workflow item. Figure~\ref{fig:task_combined_train} presents a comprehensive visualization of the task type distribution through two complementary perspectives. The left panel (bar chart) shows the workflow-level distribution by count, indicating that text-to-image generation appears in 481 workflows (54.7\% of all workflows), making it the most prevalent task type. This is followed by upscaling (348 workflows, 39.6\%), style transfer (312 workflows, 35.5\%), painting operations (228 workflows, 26.0\%), LoRA application (225 workflows, 25.6\%), and edge control (187 workflows, 21.3\%). The right panel (pie chart) displays the occurrence-level distribution, which accounts for the multi-label nature where each workflow may contain multiple task types. This perspective reveals the relative proportion of each task type among all task type occurrences, providing insight into how frequently different task types are combined within workflows.

\begin{figure*}[!tbp]
    \centering
    \includegraphics[width=1\textwidth]{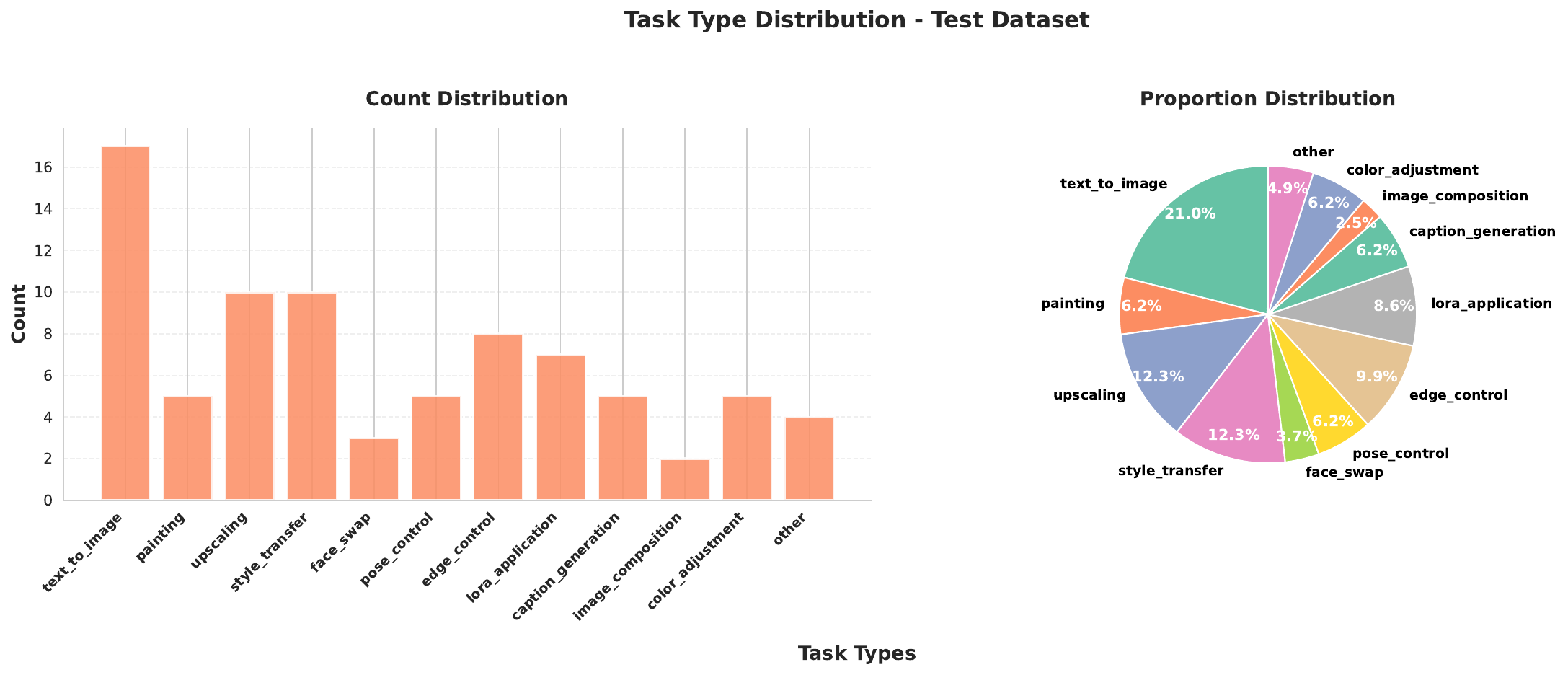}
\caption{Task type distribution in the test dataset. Left: count of workflows containing each task type (workflow-level distribution, 30 workflows total). Right: proportion of each task type among all task type occurrences (occurrence-level distribution, accounting for multi-label classification where each workflow may contain multiple task types).}
    \label{fig:task_combined_test}
\end{figure*}

\subsection{Test Dataset Analysis}

The test dataset, comprising 30 workflow items, exhibits a similar pattern with an average of 2.7 task types per item. Figure~\ref{fig:task_combined_test} illustrates the task type distribution using the same dual-perspective approach as the training dataset. The left panel (bar chart) reveals that text-to-image generation remains the most prevalent task type, appearing in 17 workflows (56.7\% of all workflows), consistent with the training dataset. Notably, the test dataset shows a higher relative proportion of edge control (8 workflows, 26.7\%) compared to the training dataset (187 workflows, 21.3\%), suggesting a focus on more controlled generation scenarios. Other significant task types in the test dataset include style transfer (10 workflows, 33.3\%), upscaling (10 workflows, 33.3\%), and LoRA application (7 workflows, 23.3\%). The right panel (pie chart) presents the occurrence-level distribution, showing how task types are distributed among all occurrences when accounting for the multi-label classification. This visualization facilitates direct comparison with the training dataset, revealing both similarities and subtle differences in task type composition between the two datasets.

\subsection{Comparative Analysis}

The comparative visualization reveals that while both datasets share similar task type distributions, the test dataset exhibits slight variations in the relative proportions of certain task types. Notably, the test dataset shows a higher emphasis on edge control (8 workflows, 26.7\%) compared to the training dataset (187 workflows, 21.3\%), and similar proportions of style transfer operations (10 workflows, 33.3\% in test vs. 312 workflows, 35.5\% in training), indicating a focus on more controlled and stylized generation scenarios. The consistent presence of multi-task workflows (averaging 2.7--2.9 task types per item) across both datasets underscores the complexity of real-world image generation pipelines, where users typically require sequential or parallel application of multiple techniques to achieve their desired results.

\paragraph{Key Findings:}

\begin{itemize}
    \item \textbf{Multi-task Nature:} The high average number of task types per workflow (2.7--2.9) confirms that real-world image generation workflows typically integrate multiple operations, such as generation, editing, and enhancement.
    
    \item \textbf{Task Type Dominance:} Text-to-image generation dominates both datasets, appearing in 481 workflows (54.7\%) in the training set and 17 workflows (56.7\%) in the test set, reflecting its fundamental role as the base operation in most image generation pipelines.
    
    \item \textbf{Post-processing Prevalence:} The high frequency of upscaling (348 workflows, 39.6\% in training; 10 workflows, 33.3\% in test) and style transfer (312 workflows, 35.5\% in training; 10 workflows, 33.3\% in test) operations indicates that users frequently apply enhancement and stylization steps after initial generation.
    
    \item \textbf{Control Mechanisms:} The significant presence of edge control (187 workflows, 21.3\% in training; 8 workflows, 26.7\% in test) and pose control operations (143 workflows, 16.3\% in training; 5 workflows, 16.7\% in test) demonstrates the importance of structural guidance in controlled image generation scenarios.
\end{itemize}

These visualizations provide comprehensive insights into the task distribution patterns, supporting our understanding of the diverse requirements and complexity inherent in modern image generation workflows.
\end{document}